\documentclass[sigconf]{acmart}

\usepackage{enumerate}
\usepackage{enumerate}
\usepackage{algorithm,algpseudocode}
\usepackage{amsmath}
\usepackage{amsfonts}
\usepackage{amssymb}

\usepackage{amssymb}
\usepackage{pifont}

\newcommand{\cmark}{\ding{51}}%
\newcommand{\xmark}{\ding{55}}%

\copyrightyear{2020}
\acmYear{2020}
\setcopyright{acmcopyright}
\acmConference[MM '20]{Proceedings of the 28th ACM International Conference on Multimedia}{October 12--16, 2020}{Seattle, WA, USA}
\acmBooktitle{Proceedings of the 28th ACM International Conference on Multimedia
(MM '20), October 12--16, 2020, Seattle, WA, USA}
\acmPrice{15.00}
\acmDOI{10.1145/3394171.3413575}
\acmISBN{978-1-4503-7988-5/20/10}

\begin{document}
\fancyhead{}

\title{HOSE-Net: Higher Order Structure Embedded Network for Scene Graph Generation}

\author{Meng Wei}
\email{weim18@mails.tsinghua.edu.cn}
\affiliation{
  \institution{Tsinghua University}
}

\author{Chun Yuan}\authornote{Chun Yuan is the corresponding author.}
\email{yuanc@sz.tsinghua.edu.cn}
\affiliation{
  \institution{Tsinghua Shenzhen International Graduate School, China, Peng Cheng Laboratory, Shenzhen, China}
}

\author{Xiaoyu Yue}
\email{yuexiaoyu002@gmail.com}
\affiliation{
}

\author{Kuo Zhong}
\email{zhongk17@mails.tsinghua.edu.cn}
\affiliation{
  \institution{Tsinghua University}
}

\renewcommand{\shortauthors}{M. Wei, et al.}

\begin{abstract}
Scene graph generation aims to produce structured representations for images, which requires to understand the relations between objects.
Due to the continuous nature of deep neural networks, the prediction of scene graphs is divided into object detection and relation classification.
However, the independent relation classes cannot separate the visual features well.
Although some methods organize the visual features into graph structures and use message passing to learn contextual information, they still suffer from drastic intra-class variations and unbalanced data distributions.
One important factor is that they learn an unstructured output space that ignores the inherent structures of scene graphs.
Accordingly, in this paper, we propose a \textbf{H}igher \textbf{O}rder \textbf{S}tructure \textbf{E}mbedded Network (HOSE-Net) to mitigate this issue. 
First, we propose a novel structure-aware embedding-to-classifier(SEC) module to 
incorporate both local and global structural information of relationships into the output space. Specifically, a set of context embeddings are learned via local graph based message passing and then mapped to a global structure based classification space. Second, since learning too many context-specific classification subspaces can suffer from data sparsity issues, we propose a hierarchical semantic aggregation(HSA) module to reduces the number of subspaces by introducing higher order structural information.
HSA is also a fast and flexible tool to automatically search a semantic object hierarchy based on relational knowledge graphs. 
Extensive experiments show that the proposed HOSE-Net achieves the state-of-the-art performance on two popular benchmarks of Visual Genome and VRD.
\end{abstract}

\ccsdesc[500]{Computing methodologies~Scene understanding}

\keywords{Scene graph generation; knowledge graph; context embedding; graph convolution network}


\maketitle

\section{Introduction}
\label{sec:intro}


\begin{figure}
    \centering
    \begin{tabular}{c c c}
    \begin{minipage}[c]{0.3\linewidth}
    \includegraphics[width=\linewidth, height=1.4\linewidth]{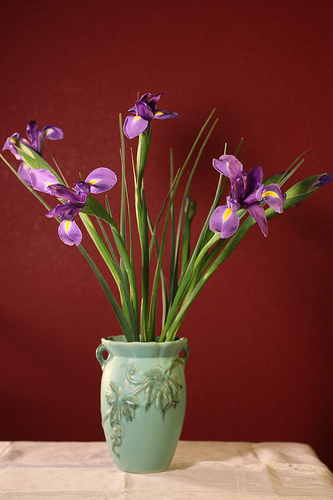}
    \end{minipage} & 
    \begin{minipage}[c]{0.3\linewidth}
    \includegraphics[width=\linewidth, height=1.4\linewidth]{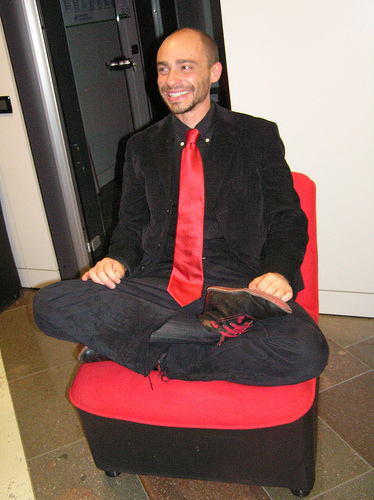}
    \end{minipage} &
    \begin{minipage}[c]{0.3\linewidth}
    \includegraphics[width=\linewidth, height=1.4\linewidth]{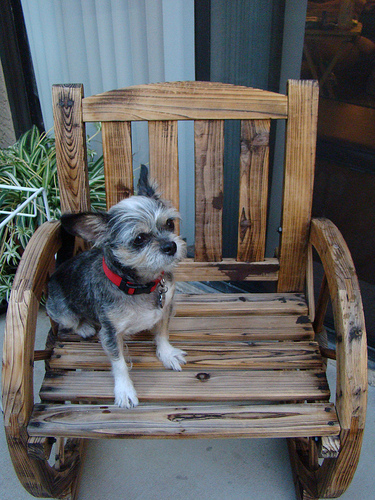}
    \end{minipage} \\
    (a) & (b) & (c) \\ 
    \multicolumn{3}{c}{\includegraphics[width=\linewidth]{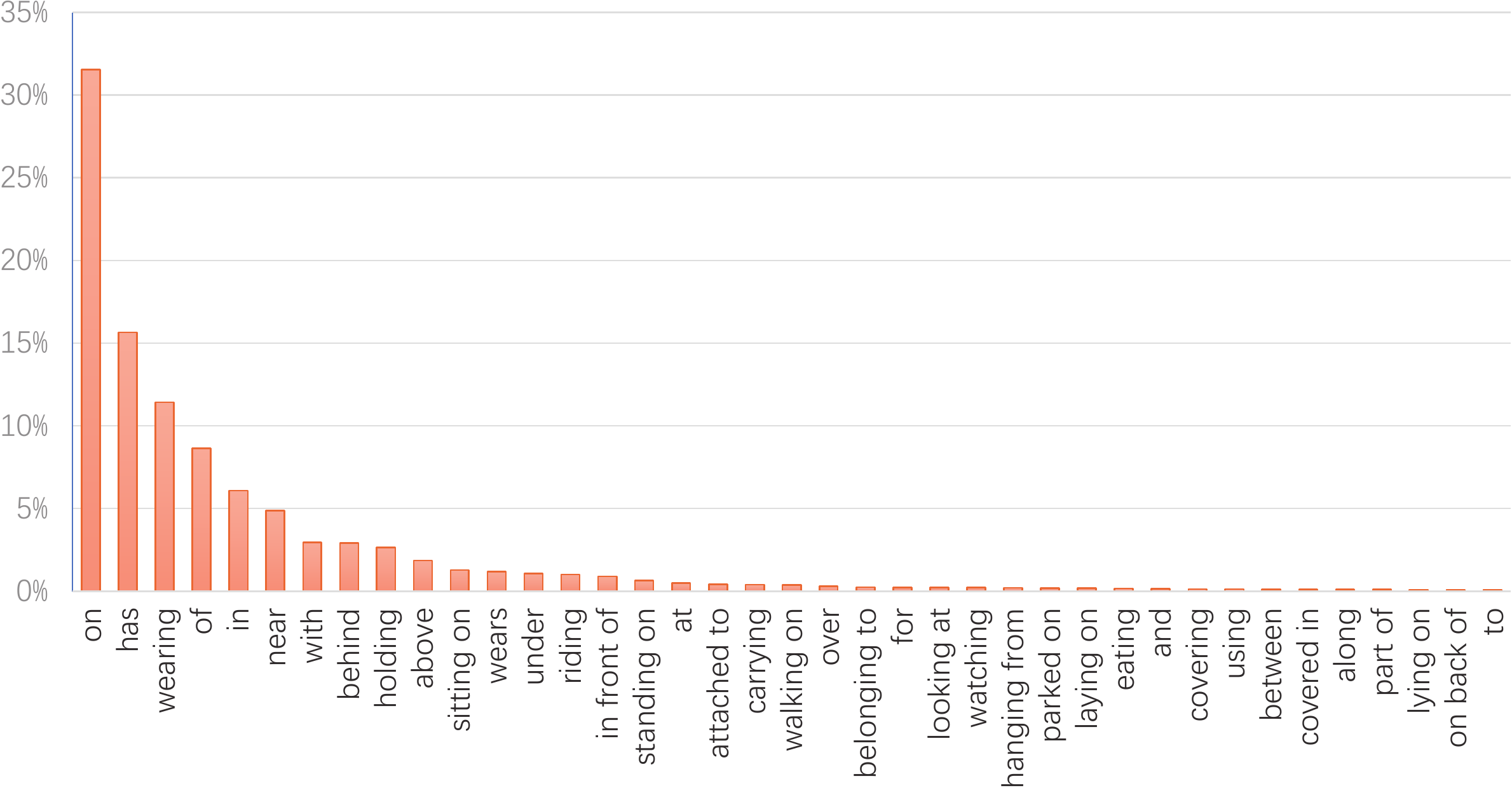}} \\
    \multicolumn{3}{c}{(d)}\\
    \end{tabular}
    \caption{(a): $\textless$vase-sitting on-table$\textgreater$; (b): $\textless$man-sitting on-chair$\textgreater$; (c): $\textless$dog-sitting on-chair$\textgreater$. (a)(b)(c) have completely different visual appearances but are considered as the same relation class. (d): The long-tailed distribution of independent relation classes}
    \label{fig:motivation}
\end{figure}

\begin{figure*}
    \centering
    \begin{tabular}{c c c}
    \begin{minipage}[c]{0.36\linewidth}
    \includegraphics[width=\linewidth]{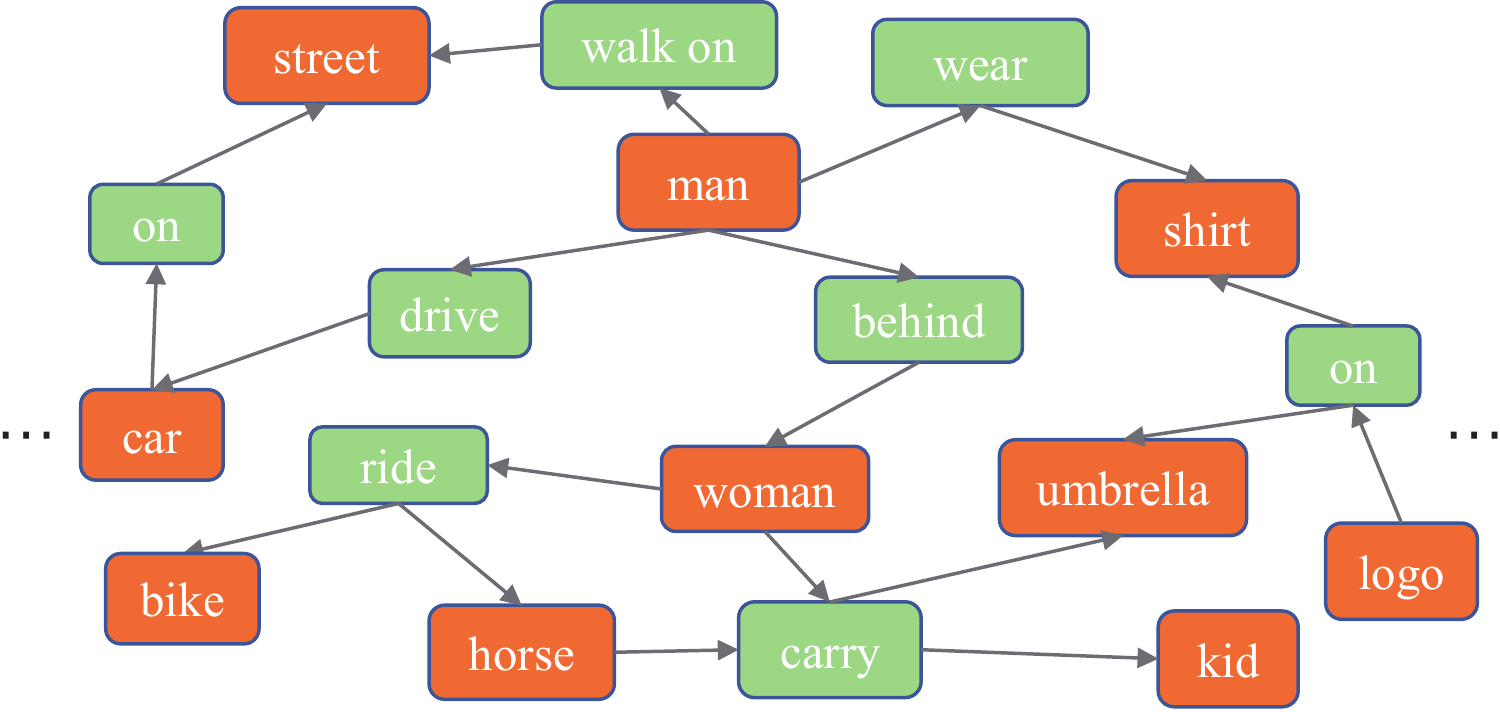}
    \end{minipage} &
    \begin{minipage}[c]{0.23\linewidth}
    \includegraphics[width=\linewidth]{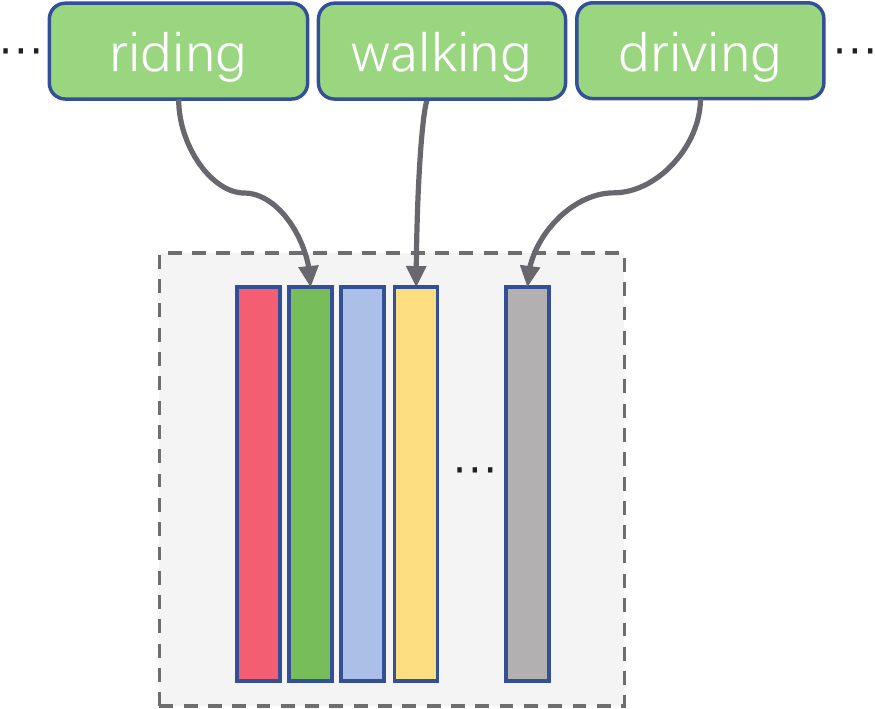}
    \end{minipage} & 
    \begin{minipage}[c]{0.36\linewidth}
    \includegraphics[width=\linewidth]{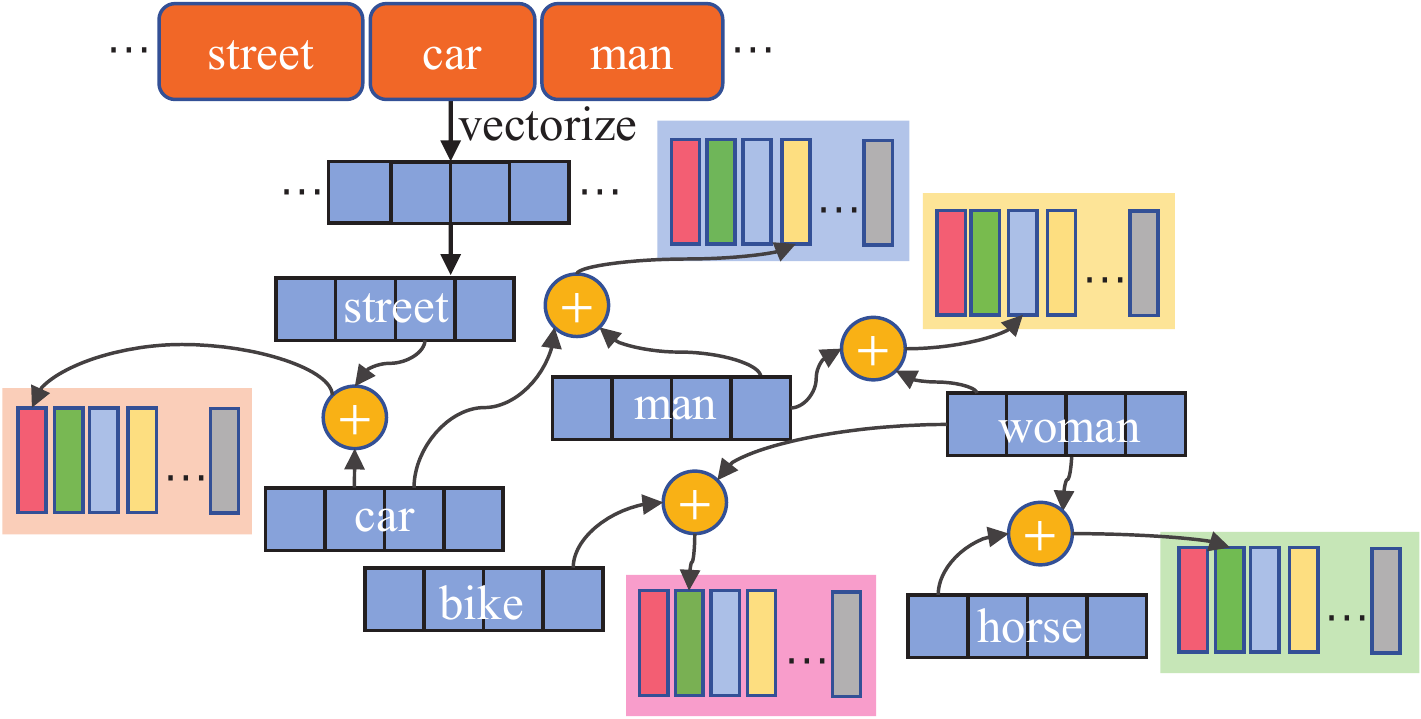}
    \end{minipage} \\
    (a) & (b) & (c) \\ 
    \end{tabular}
    \caption{(a): The global knowledge graph of VG; (b): Unstructured output space in which the relation classifier is shared among all subject-object pairs; (c): Structured output space of HOSE-Net in which the relation classifier is context-specific.}
    \label{fig:intro}
\end{figure*}
In recent years, visual recognition tasks for scene understanding has gained remarkable progress, particularly in object detection and instance segmentation. While accurate identification of objects is a critical part of visual recognition, higher-level scene understanding requires higher-level information of objects. 
Scene graph generation aims to provide more comprehensive visual clues than individual object detectors by understanding object interactions. Such scene graphs serve as structural representations of images by describing objects as nodes (``subjects/objects") and their interactions as edges (``relation"), which benefit many high-level vision tasks such as image caption\cite{li2019know,yang2019auto,guo2019aligning}, visual question answering\cite{teney2017graph,peng2019cra} and image generation\cite{johnson2018image}. 

In scene graph generation, we actually obtain a set of visual phases$\textless$ subject-relation-object $\textgreater$ and the locations of objects in the image.
The triples of each scene graph form a local knowledge graph of the image and the triples of the whole training set form a global knowledge graph of relationships as shown in Figure~\ref{fig:intro} (a).
It remains a challenging task because deep neural networks cannot directly predict structured data due to its continuous nature.
It's a common practice to divide the scene graphs into classifiable graph elements.
\cite{sadeghi2011recognition} divides them into visual phase classes.
However, it's infeasible due to the hyper-linear growth concerning the number of objects and relations.
A widely-adopted strategy is to divide them into independent object classes and relation classes\cite{lu2016visual}.
Most methods classify the objects separately and then apply local graph structures to learn contextual object representations for relation classification\cite{xu2017scene,qi2019attentive,chen2019knowledge}.
However, they ignore the fact that the output space should also be contextual and structure-aware and adopt an unstructured one as shown in Figure~\ref{fig:intro} (b).  
Hence these methods suffer from drastic intra-class variations.
For example, given the relation ``sit on'', the visual contents vary from ``vase-sit on-table" to ``dog-sit on-chair" as shown in Figure~\ref{fig:motivation} (a)(b)(c).
On the other hand, the distribution of these independent relation classes is seriously unbalanced as shown in Figure~\ref{fig:motivation} (d).

To mitigate the issues mentioned above, we propose a novel higher order structure embedded network (HOSE-Net), which consists of a visual module, a structure-aware embedding-to-classifier (SEC) module and a hierarchical semantic aggregation(HSA) module.
The SEC module is designed to construct a contextual and structured output space.
First, since objects serve as contexts in relationships, SEC learns context embeddings which embeds the objects' behavior patterns as subjects or objects and transfers this knowledge among the classifiers it connects to based on the overall class structure.
It adopts a graph convolution network\cite{kipf2016semi} to propagate messages on the local graphs with the guidance of object co-occurrence\cite{mensink2014costa} statistics. 
Second, SEC learns a mapping function to project the context embeddings to related relation classifiers.
This mapping function is shared among all contexts which implicitly encodes the statistical correlations among objects and relations and organize a global knowledge graph based output space shown in Figure~\ref{fig:intro} (c). 
Since the unbalanced relation data are distributed into different subspaces, SEC can alleviate the long-tailed distribution and the intra-class variations.

However, even if the context-specific classifiers can share statistical strengths via the context embeddings, distributing the training samples into a large set of subspaces can still suffer from sparsity issues.
To address this problem, we are inspired by the thought that object-based contexts can be redundant or noisy since relations are often defined in more abstract contexts. 
For example, ride in ``man-ride-horse'' and ``woman-ride-elephant'' can be summarized as ``people-ride-animal''.
Accordingly, we propose a hierarchical semantic aggregation (HSA) module to mine the latent higher order structures in the global knowledge graph.
HSA hierarchically clusters the graph nodes following the principle that, if two objects have similar behavior patterns in the relationships they involved, the contexts they create can be embedded together, which is designed to find a good strategy to redistribute the samples into a smaller set of subspaces.
An object semantic hierarchy is generated in the process even if HSA just uses the graph structures.
It's not hard to understand because a semantic hierarchy is based on the properties of objects which also very relevant to their behavior patterns in relationships. 

In summary, the proposed Higher Order Structure Embedded Network(HOSE-Net) uses embedding methods to construct a structured output space.
By modeling the inter-dependencies among object labels and relation labels, the serious intra-class variations and the long-tailed distribution can be alleviated. Moreover, clustering methods are used to make the structured output space more scalable and generalized.

\begin{figure*}
    \centering
    \includegraphics[width=\linewidth]{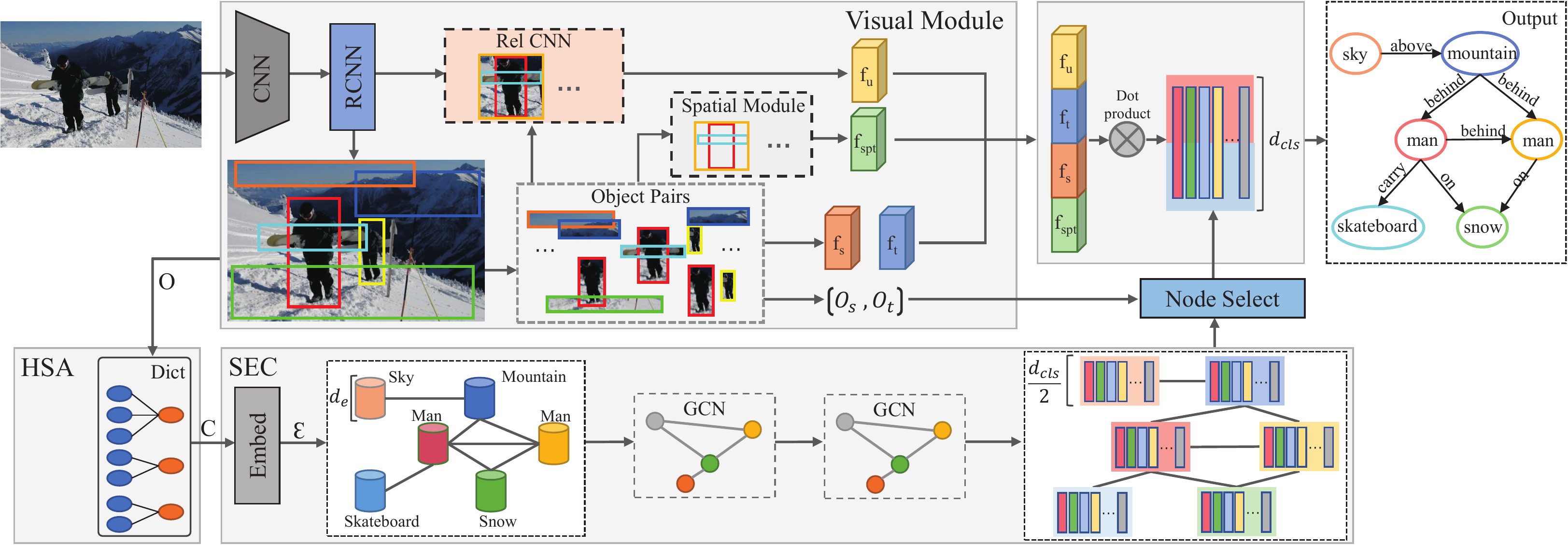}
    \caption{The framework of our HOSE-Net. It consists of three modules: (1) a visual module which outputs detection results and prepare subject-object pairs for relation representation learning;
    (2) a SEC module which embeds the object labels into context embeddings by message passing and maps them to classifiers;
    (3) a HSA module which distill higher order structural information for context embedding learning.}
    \label{fig:overview}
\end{figure*}

Our contributions are as follows:
\begin{enumerate}[(1)]
    \item We propose to map the object-based contexts in relationships into a high-dimensional space to learn contextual and structure-aware relation classifiers via a novel structure-aware embedding-to-classifier module. This module can be integrated with other works focusing on visual feature learning.
    \item We design a hierarchical semantic aggregation module to distill higher order structural information for learning a higher order structured output space. 
    \item We extensively evaluate our proposed HOSE-Net, which achieves new state-of-the-art performances on challenging Visual Genome \cite{krishna2017visual} and VRD~\cite{lu2016visual} benchmarks.
\end{enumerate}


\section{Related Work}
\textbf{Scene Graph Generation.}
Recently, the task of scene graph generation is proposed to understand the interactions between objects. 
\cite{sadeghi2011recognition} decomposes the scene graphs into a set of visual phase classes and designs a detection model to directly detect them from the image.
Considering each visual phase as a distinct class would fail since the number of visual phase triples can be very large even with a moderate number of objects and relations.
An alternative strategy is to decompose the scene graphs into object classes and relation classes in which way the graph structures of the output data is completely collapsed.
Most of these methods focus on modeling the inter-dependencies of objects and relations in the visual representation learning.
\cite{dai2017detecting} embeds the statistical inference procedure into the deep neural network via structural learning.
\cite{xu2017scene} constructs bipartite sub-graphs of scene graphs and use RNNs to refine the visual features by iterative message passing.
\cite{chen2019knowledge,qi2019attentive,cui2018context} uses graph neural networks to learn contextual visual representation.
However, these methods still suffer from highly diverse appearances within each relation class because they all adopt a flat and independent relation classifiers.
In this paper, we argue that the structural information including the local and the global graph structures of the output data is vital for regularizing a semantic space.
\\
\textbf{Learning Correlated Classifiers With Knowledge Graph}. 
Zero-shot learning(ZSL) models need to transfer the knowledge learned from the training classes to the unknown test classes.
A main research direction is to represent each class with learned vector representations.
In this way, the correlations between known classes and unknown classes can help to transfer the knowledge learned from the training classes to the unknown test classes by mapping the embeddings to visual classifiers.
Knowledge Graphs (KGs) effectively capture explicit relational knowledge about individual entities hence many methods\cite{wang2018zero,kampffmeyer2019rethinking,hascoet2019semantic,gao2019know,zhang2019tgg} use KGs to learn the class correlations.
In scene graph generation, the relation classes are correlated by object classes as in the knowledge graph and the structural information is vital for a well-defined output space.
We indirectly learn vector representations of the objects' role in relationships which are mapped to the visual relation classifiers via the knowledge graph structure.

\section{Approach}

\subsection{Overview}
We formally define a scene graph as $G = \left \{ B, O, R \right \}$. $O = \left \{ o_1, o_2, \dots, o_n \right \}$ is the object set and $o_i$ denotes the i-th object in image. $B = \left \{ b_1, b_2, \dots, b_n \right \}$ is the bounding box set and $b_i$ denotes the bounding box of $o_i$. $R = \left \{ r_{o_1 \rightarrow o_2}, r_{o_2 \rightarrow o_3}, \dots, r_{o_{(n-1)} \rightarrow o_n} \right \}$ is the edge set and $r_{o_i \rightarrow o_j}$ denotes the relation between subject $o_i$ and object $o_j$. 
The probability distribution of the scene graph $\Pr(G | I)$ is formularized as:
\begin{equation}
    \Pr(G | I) = \Pr(B|I)\Pr(O|B,I)\Pr(R|B,O,I)
\end{equation}
We follow the widely-adopted two-stage pipeline\cite{zellers2018neural} to generate scene graphs. The first stage is object detection including object localization ($\Pr(B|I)$) and object recognition ($Pr(O|B, I)$). The second stage is relation classification ($\Pr(R|B,O,I)$). 
Our proposed HOSE-Net consists of a visual module, a SEC module and a HSA module.

Section ~\ref{sec:visrep} introduces the visual module.
The major component is an object detector, 
which outputs $B$, $O$ and the region features $F=\left \{ f_1, f_2, \dots, f_n \right \}$.
Then a set of object pairs $\left \{ (f_s, f_t), (o_s, o_t), (b_s,b_t) \right \}$ are produced, where $ s \neq t;s, t = 1 ... n $.
The union box feature $f_u$ for each pair is extracted by a relation branch.
The spatial feature $f_{spt}$ for each pair is learned from $(b_s,b_t)$ by a spatial module.
Section ~\ref{sec:secm} introduces the structure-aware embedding-to-classifier(SEC) module.
First, we construct local graphs to transfer statistical information between context embeddings of $O$.
Then the context embeddings are mapped to a set of primitive classifiers.
The classifier for each relation representation is adaptively generated by concatenating the primitive classifiers according to the pair label $(o_s, o_t)$.
Section ~\ref{sec:hsa} introduces the hierarchical semantic aggregation(HSA) module.
Based on the resulting semantic hierarchy, HSA creates a context dictionary $\mathcal{D}$ to map $o_i$ to one-hot encoding $c_i \in \mathbb{R}^{K}$ where $K  \in \left [ 1,N \right ] $ is the number of context embeddings and $N$ is the number of object classes. 
The overall pipeline in shown in Figure~\ref{fig:overview}.

\subsection{Visual Representation}\label{sec:visrep}
\textbf{Object Detection}. In the first stage, the object detection is implemented by a Faster RCNN\cite{ren2015faster}. With the detection results, a set of subject-object region feature pairs $(f_s, f_t)$ with label information $(o_s, o_t)$ and coordinates of subject box $(x_s,y_s,w_s,h_s)$, object box $(x_t,y_t,w_t,h_t)$, union box$(x_u,y_u,w_u,h_u)$ is produced.
Then a separate relation branch uses three bottlenecks to refine the image feature and extract the union box feature $f_u$ of each subject-object pair by roi pooling.
While the Faster RCNN branch focuses on learning discriminative features for objects, the relation branch focuses on learning interactive parts of two objects.\\
\\
\textbf{Relation Representation}. Most existing methods explore the the visual representation learning for relations. To establish the connections between objects, they usually build graphs to associate the detected regions and use message passing frameworks to learn contextualized visual representations. Then the fusion features of the subjects and objects are projected to a set of independent relation labels by a softmax function.
Whether the relation classifiers are structured and contextualized has been little explored.
To verify the effectiveness of adopting a structured output space, we don't use a graph structure for learning the visual representations.
Given the triple region features from the detection module $(f_s, f_t, f_u)$, the visual representation of the relation is:
\begin{equation}
    r_{st} = \Psi_{st}([f_u; f_s; f_t])
\end{equation}
where $[;]$ is the concatenation operation and $\Psi_{st}$ is a linear transformation. $[f_u, f_s, f_t] \in \mathbb{R}^{3d_f}$. \\
\\
\textbf{Spatial Representation}.
The relative positions of the subject boxes and the object boxes are also valuable spatical clues for recognizing the relations.
The normalized box coordinates $\widehat{b_i}$ are computed as $[\frac{x}{w_{img}},\frac{y}{h_{img}},\frac{x+w}{w_{img}},\frac{y+h}{h_{img}},\frac{wh}{w_{img} h_{img}}]$ where $w_{img}$ and $h_{img}$ are the width and height of the image. 
The relative spatial feature $b_{st}$ is encoded as $[\frac{x_s-x_t}{w_t},\frac{y_s-y_t}{h_t},log\frac{w_s}{w_t},log\frac{h_s}{h_t}]$. The final spatial representation is the concatenation of the normalized features and the relative features of the subject and object boxes:
\begin{equation}
    f_{spt} = \Psi_{spt}([\widehat{b_s},\widehat{b_t},b_{st}])
\end{equation}
where $\Psi_{spt}$ is a linear transformation, $[\widehat{b_s},\widehat{b_t},b_{st}] \in \mathbb{R}^{14}$.

\subsection{Structure-Aware Embedding-to-Classifier}\label{sec:secm}
Given the object label information $O = \left \{ o_1, o_2, \dots, o_n \right \}$, our proposed SEC module generates dynamic classifiers for relation representations according to the pair label $(o_s,o_t)$.
First, we embed the object labels into higher level context embeddings. 
The one-hot context encodings of objects $\mathcal{C} = \left \{ c_1, c_2, \dots, c_n \right \},c_i \in \mathbb{R}^{K}$ are obtained through the context dictionary $\mathcal{D}$ which will be discussed in \ref{sec:hsa}. 
The context embeddings $\mathcal{E} = \left \{ e_1, e_2, \dots, e_n \right \}, e_i \in \mathbb{R}^{d_e}$ are genereted as follows:
\begin{equation}
    e_i = W_{e}c_i
\end{equation}
where $W_{e} \in \mathbb{R}^{d_e \times K}$ is a context embedding matrix to be learned.

Then the context embeddings $\mathcal{E}$ are fed into a graph convolution network to learn local contextual information based on object co-occurrences.
We model the co-occurrence pattern in the form of conditional probability, i.e., $\mathcal{P}_{ij}$, which denotes the probability of occurrence of the j-th object class when the i-th object class appears. We compute the co-occurrences of object pairs in the training set and get the matrix $\mathcal{T}\in \mathbb{R}^{N \times N}$, $N$ is the number of object classes. $\mathcal{T}_{ij}$ denotes the co-occurring times of label pairs. The conditional probability matrix $\mathcal{P}$ is computed by:
\begin{equation}
    \mathcal{P}_{ij} = \frac{\mathcal{T}_{ij}}{\sum_{j}^{N}\mathcal{T}_{ij}}
\end{equation}
where $\sum_{j}^{N}\mathcal{T}_{ij}$ denotes the total number of i-th object class occurrences in the training set.
Then the adjacency matrix $A \in \mathbb{R}^{n \times n}$ of the local contextual graph is produced by:
\begin{equation}
    A_{ij} = \mathcal{P}_{o_i, o_j}
\end{equation}
The update rule of each GCN layer is:
\begin{equation}
    \mathcal{E} ^{l + 1} = f(\mathcal{E} ^{l}, A)
\end{equation}where $f$ is the graph convolution operation of \cite{kipf2016semi}.
The node output of the final GCN layer is the primitive classifiers $W_{prim} = \left \{ w_1, w_2, \dots, w_n \right \}, w_i \in \mathbb{R}^{\frac{d_{cls}}{2}}$ formulated as:
\begin{equation}
    w_i = {e_i}^{l + 1} = \sigma(\sum_{j\in \mathbb{N}_{i}}A_{ij}U{e_j}^{l}) 
\end{equation}
where $U \in \mathbb{R}^{d_e \times \frac{d_{cls}}{2}}$ is the transformation matrix to be learned. $\mathbb{N}_{i}$ is the neighbor node set of $e_i$. $\sigma$ is the nonlinear function.
For each subject-object label pair $(o_s,o_t)$, the visual classifier $W_{st}$ is a composition of two primitive classifiers according to its context:
\begin{equation}
    W_{st} = [w_{o_s};w_{o_t}] \in \mathbb{R}^{d_{cls}}
\end{equation}
where $[;]$ is the concatenation operation.
Apply the learned classifier to the relation representations to get the predicted scores:
\begin{equation}
    \hat{y} = W_{st}[r_{st};f_{spt}]
\end{equation}

\subsection{Hierarchical Semantic Aggregation}\label{sec:hsa}

\begin{figure}
    \centering
    \begin{minipage}[c]{\linewidth}
    \includegraphics[width=\linewidth]{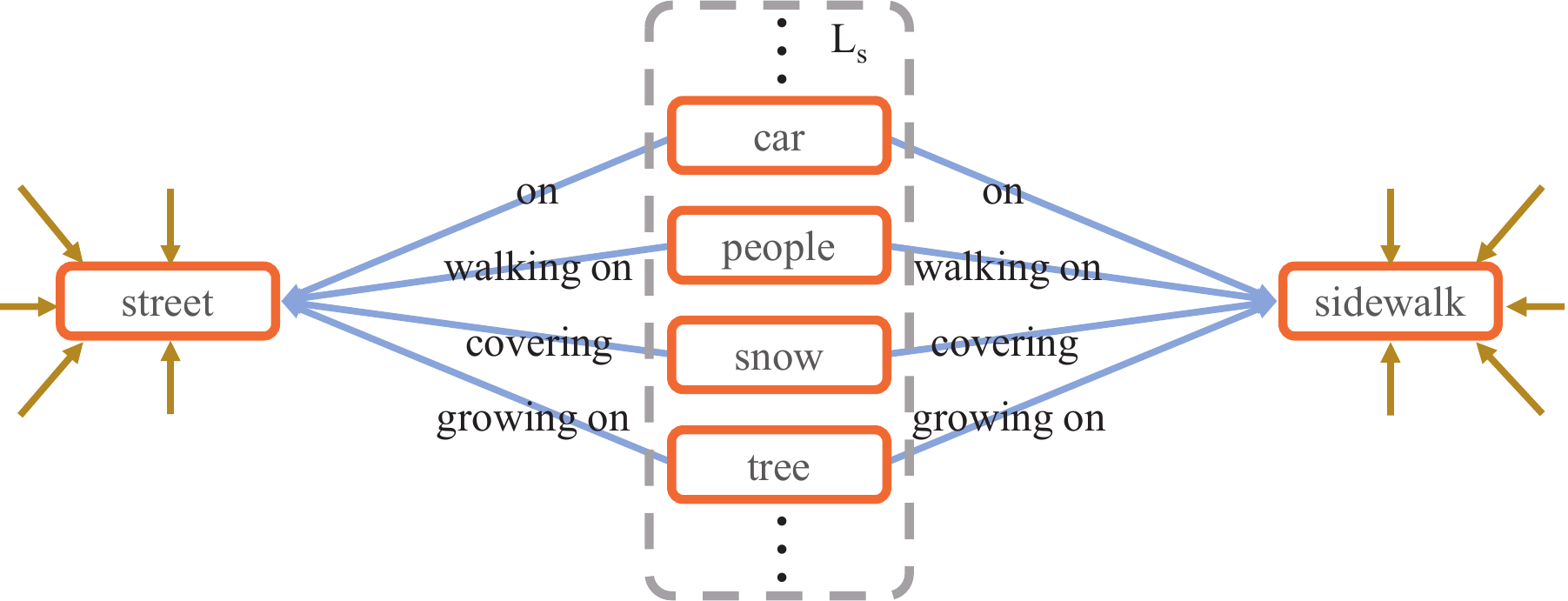}
    \end{minipage}
    (a)
    \begin{minipage}[c]{\linewidth}
    \includegraphics[width=\linewidth]{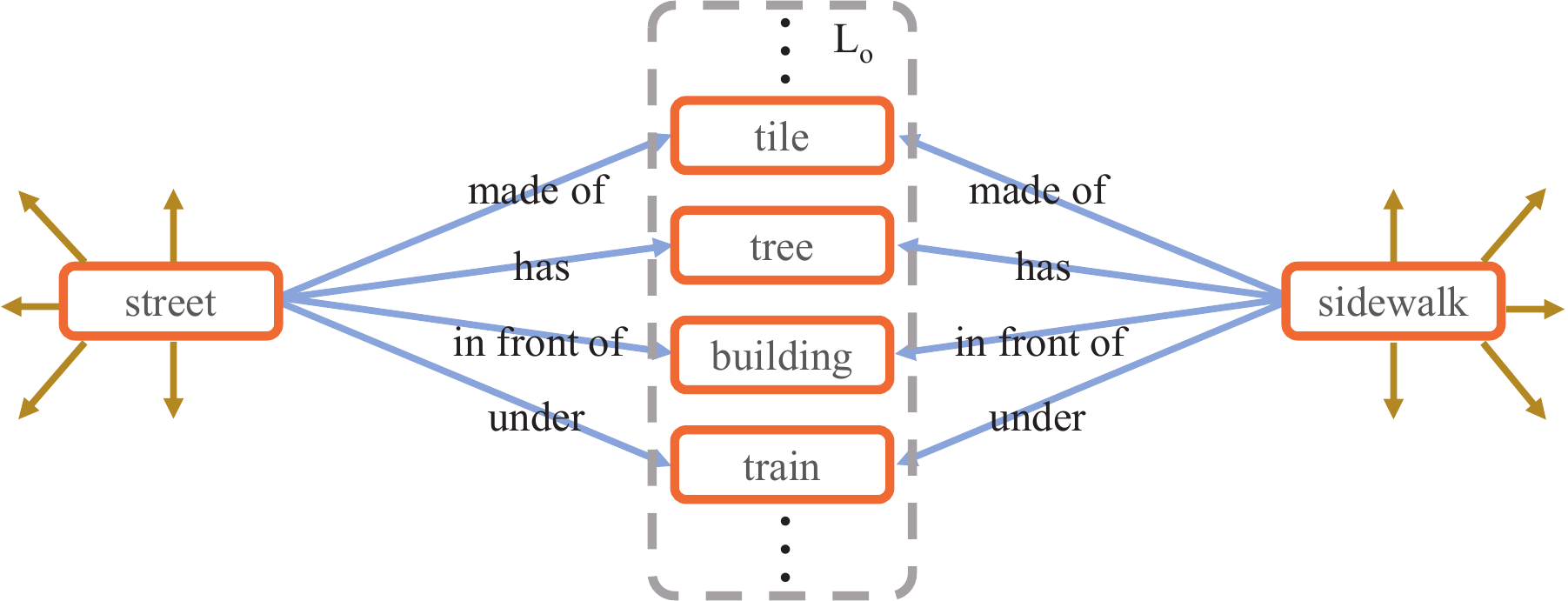}
    \end{minipage}
    (b)
    \caption{The connectivity subgraph of (street, sidewalk) is illustrated in two parts: (street, sidewalk) are objects (a) or subjects (b) in $\textless$subject-relation-object$\textgreater$. The blue edges are the connected path and 
    the yellow edges are the unconnected path.}
    \label{fig:scene_graph_example}
\end{figure}
Even if the global knowledge graph exhibits rich, lower-order connectivity patterns captured at the level of objects and relations, new problems emerge if we create context embeddings for all object classes. When the number of classes $N$ increases, a lot of context-specific classifiers can't get sufficient training samples due to data sparsity and can not be scalable.
The motivation of HSA is, although relations exist among concrete objects, the objects actually have many similar higher level behavior patterns in the overall contexts.
And there exists higher order connectivity patterns on the class structure which are essential for understanding the object behaviors in relationships.
%

We design an clustering algorithm for mining the higher order structural information based on behavior patterns of objects.
The connectivity pattern with respect to two nodes $q_s, q_t$ of knowledge graph $KG$ is represented in a subgraph $SG$.
$SG$ includes two sets of connection nodes $L_s,L_o$ between $q_s, q_t$: 
\begin{eqnarray}
q_i\in L_s \Leftrightarrow r_{q_i \rightarrow q_s} = r_{q_i \rightarrow q_t}\\
q_i\in L_o \Leftrightarrow r_{q_i \leftarrow q_s} = r_{q_i \leftarrow q_t}
\end{eqnarray}
where $r_{q_i \rightarrow q_j}$ denotes the relation between $q_i$ and $q_j$.
Figure~\ref{fig:scene_graph_example} illustrates an example, which visualize the common patterns between street and sidewalk: both of them are made of tile, can be covered in snow, a person can walk on them and have buildings nearby.

If the behavior patterns of $q_s$ and $q_t$ have a large overlap, they can be clustered into a higher-level node.
The similarity score between $q_s, q_t$ is defined as:
\begin{equation}
\begin{split}
    f_{sim}(q_s, q_t) & = \frac{\left | L_s \right |}{d_{in}(q_s) + d_{in}(q_t) - \left | L_s \right |} \\
     & +  \frac{\left | L_o \right |}{d_{out}(q_s) + d_{out}(q_t) - \left | L_o \right |},
\end{split}
\end{equation}
where $d_{in}(q_i)/d_{out}(q_i)$ denotes the number of incoming/outgoing edges of node $q_i$, which represents the occurrence times of $q_i$ in all relationships as object/subject respectively. $\left | L_s \right|/\left | L_o \right |$ denotes the number of nodes in $L_s/L_o$, which represents the number of common behavior patterns of $q_s, q_t$ as object/subject.
This measure is fully based on the graph structure, not on the distributed representations of nodes from external knowledge graphs.
\\
\begin{algorithm}[tb]
\caption{}
\label{alg:optim}
\begin{algorithmic}[1] 
\Require{$KG = \left \{ (q_1^0,q_2^0,...,q_N^0),(r_{q_1^0 \rightarrow q_2^0},...,r_{q_i^0 \rightarrow q_j^0},...) \right \}$,similarity measure function $f_{sim}$, cluster number $K$} 
    \For{\texttt{$i = 1,2,...,N$}}
        \State{$\lambda_{i} = 1 $}
        \For{\texttt{$j = 1,2,...,N$}}
            \State{$Sim(i,j) = f_{sim}(q_i^0,q_j^0)$}
            \State{$Sim(j,i) = Sim(i,j)$}
        \EndFor
    \EndFor
    \State {Set current cluster number $num = N$}
    \While{$num > K$}\Comment{Find the two most similar node cluster}
        \State{ $q_{i}^{l_i}, q_{j}^{l_j} \gets SELECTMAX(Sim(i,j) / (\lambda_{i} + \lambda_{j}))$}
        \State $q_{i}^{l_{ij}} \gets MERGE(q_{i}^{l_i},q_{j}^{l_j})$
        \State $KG,Sim \gets UPDATE(KG,Sim)$
        \State $\lambda_{i} \gets \lambda{i} + \lambda{j} + 1$
        \State $REINDEX(\lambda)$
        \State $num \gets num - 1$
    \EndWhile
    \State{$\mathcal{D} \gets GETDICT((q_1^0,q_2^0,...,q_N^0),(q_1^l,q_2^l,...,q_K^l))$}
\Ensure{$\mathcal{D}$}
\end{algorithmic}
\end{algorithm}

\noindent\textbf{Algorithm.} We use hierarchical agglomerative clustering to find the node clusters shown in Algorithm~\ref{alg:optim}.
At each iteration, we merge the two clusters which have the most similar behavior patterns and update the knowledge graph by replacing the two clustered nodes with a higher level node.
Since the given triples are incomplete and unbalanced, we introduce a penalty term $\lambda$ to avoid the objects which have frequent occurrences in annotated relationships dominating the clustering.
When the number of clusters reaches the given K, the algorithm stops iterating.
We encode the clustering results as a dictionary $\mathcal{D}$ to map the N objects to one-hot encodings of dimension $K$ hence the objects within the same cluster will have the same context embedding.
In this way, the output space is reorganized into a smaller one.
Even if the clusters are not reasonable for all relations, experiments show that the context embeddings can still learn the upside in a high-dimensional space.
\\

\begin{figure*}
    \centering
    \begin{tabular}{c c}
    \begin{minipage}[c]{0.48\linewidth}
    \includegraphics[height=\linewidth]{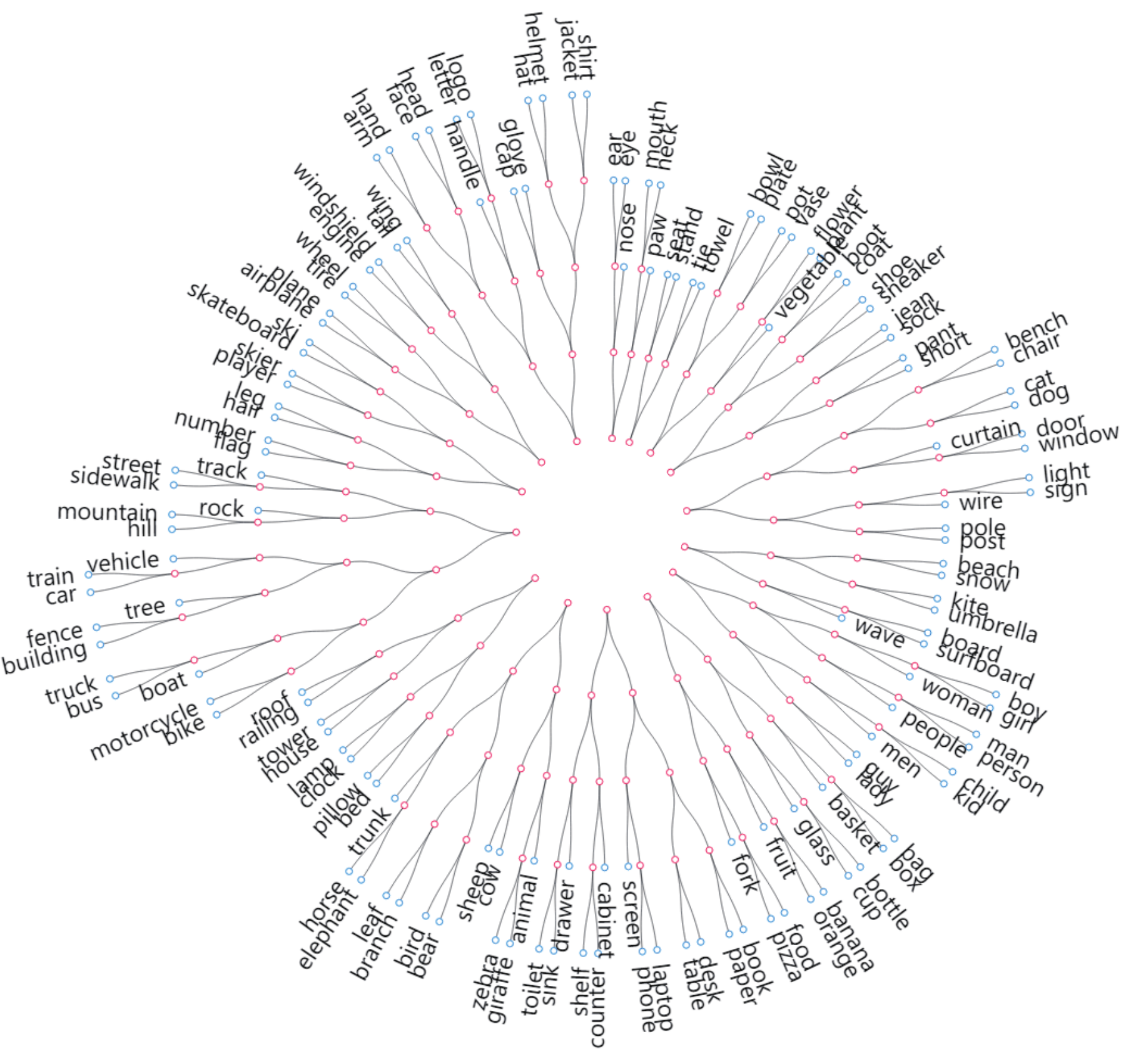}
    \end{minipage} &
    \begin{minipage}[c]{0.52\linewidth}
    \includegraphics[height=\linewidth]{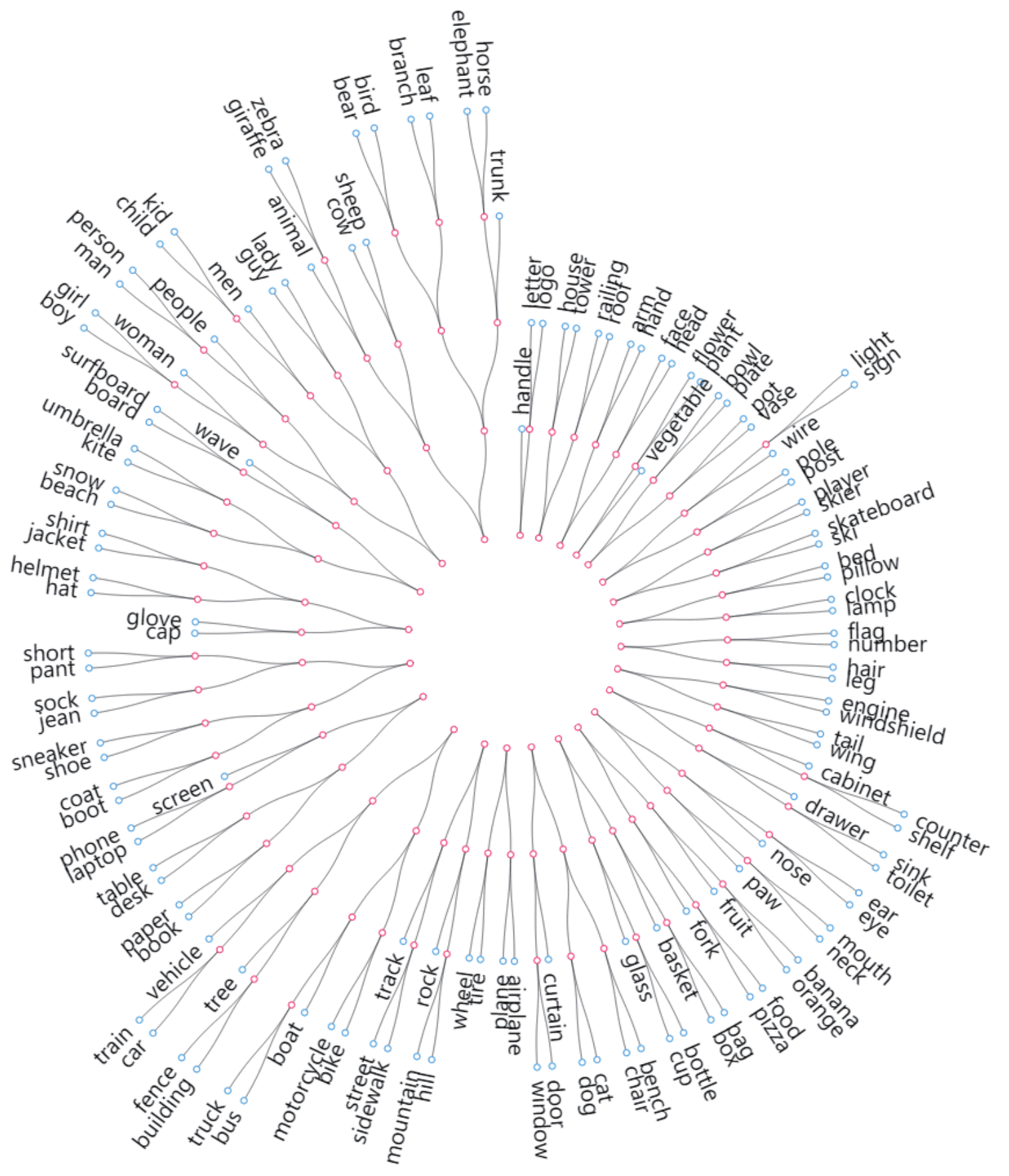}
    \end{minipage} \\
    (a) & (b) \\ 
    \end{tabular}
    \caption{(a) The semantic hierarchy of VG. (b) The semantic hierarchy of VRD.}
    \label{fig:optim_result}
\end{figure*}
\section{Experiments}

\begin{table*}
    \centering
    \begin{tabular}{l l | l | l l | l l | l l}
    \hline
SEC & HSA & & \multicolumn{2}{c}{SGDET} \vline & \multicolumn{2}{c}{SGCLS} \vline & \multicolumn{2}{c}{PRDCLS}\\
 & & Recall at & 50 & 100 & 50 & 100 & 50 & 100 \\
\hline
\xmark & \xmark & Baseline $(K=1)$ & 28.1 & 32.5 & 34.8 & 36.4 & 64.6 & 67.3\\
\cmark & \xmark & HOSE-Net $(K=150)$ & 28.6 & 33.1 & 36.2 & 37.3 & 66.5 & 69.0\\
\cmark & \cmark & HOSE-Net $(K=40)$ & \textbf{28.9} & \textbf{33.3} & \textbf{36.3} & \textbf{37.4} & \textbf{66.8} & \textbf{69.2}\\
\hline
    \end{tabular}
    \caption{Ablation study on the SEC module and HSA module.}
    \label{tab:compair_m}
\end{table*}

\begin{figure}
    \centering
    \includegraphics[width=\linewidth, height=0.45\linewidth]{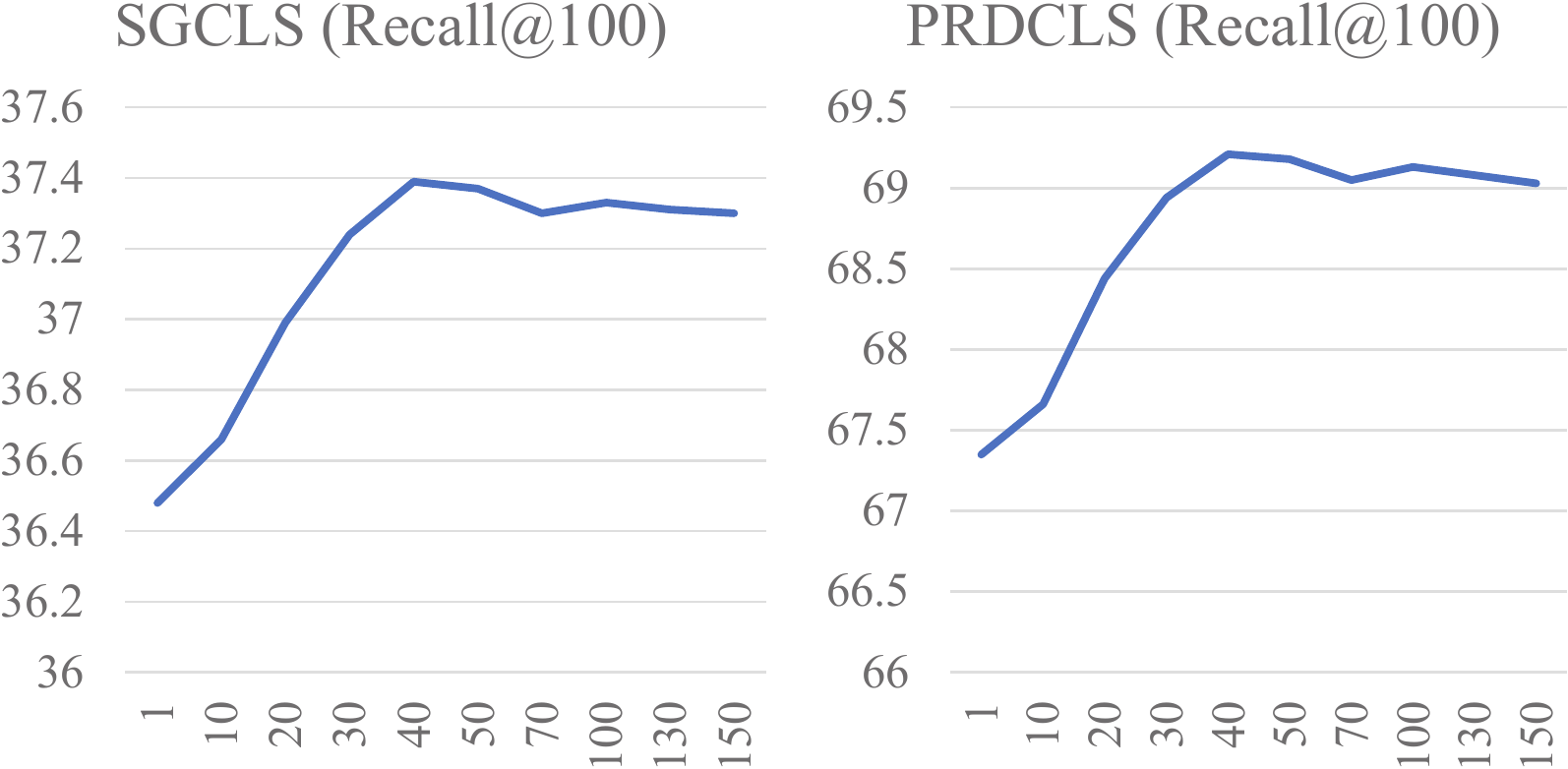}
    \caption{Ablation study on the clustering number. We draw the performance curves of SGCLS and PRDCLS on K = 1, 10, 20, 30, 40, 50, 70, 100, 130, 150.}
    \label{fig:compair_m}
\end{figure}

\begin{table*}
    \centering
    \begin{tabular}{l | l l | l l | l l}
    \hline
 & \multicolumn{2}{c}{SGDET} \vline & \multicolumn{2}{c}{SGCLS} \vline & \multicolumn{2}{c}{PRDCLS}\\
 Recall at & 50 & 100 & 50 & 100 & 50 & 100 \\
\hline
SEC ($K=150$) & 28.6 & 33.1 & 36.2 & 37.3 & 66.5 & 69.0\\
SEC + kmeans with word2vec embedding($K=40$) & 28.7 & 33.1 & 36.2 & 37.3 & 66.4 & 68.9\\
SEC + HSA ($K=40$) & \textbf{28.9} & \textbf{33.3} & \textbf{36.3} & \textbf{37.4} & \textbf{66.8} & \textbf{69.2}\\
\hline
    \end{tabular}
    \caption{ Comparison between the clustering in HSA module and kmeans with word2vec embedding clustering.}
    \label{tab:kmeans}
\end{table*}


\begin{table*}  
\centering
\begin{tabular}{l | l l l l l l | l l  l l  l l}
\hline
 & \multicolumn{6}{c}{Graph Constraint}  \vline & \multicolumn{6}{c}{No Graph Constraint}\\
 & \multicolumn{2}{c}{SGDET}  & \multicolumn{2}{c}{SGCLS}  & \multicolumn{2}{c}{PRDCLS} \vline &  \multicolumn{2}{c}{SGDET}  & \multicolumn{2}{c}{SGCLS}  & \multicolumn{2}{c}{PRDCLS}\\
Recall at & 50 & 100 & 50 & 100 & 50 & 100 & 50 & 100 & 50 & 100 & 50 & 100 \\
\hline
VRD~\cite{lu2016visual} & 0.3 & 0.5 & 11.8 & 14.1 & 27.9 & 35.0 & - & - & - & - & - & -\\
ISGG~\cite{xu2017scene} & 3.4 & 4.2 & 21.7 & 24.4 & 44.8 & 53.1 & - & - & - & - & - & -\\
MSDN~\cite{xu2017scene} & 7.0 & 9.1 & 27.6 & 29.9 & 53.2 & 57.9 & - & - & - & - & - & - \\
AsscEmbed~\cite{newell2017pixels} & 8.1 & 8.2 & 21.8 & 22.6 & 54.1 & 55.4 & 9.7 & 11.3 & 26.5 & 30.0 & 68.0 & 75.2\\
Message Passing+~\cite{zellers2018neural} & 20.7 & 24.5 & 34.6 & 35.4 & 59.3 & 61.3 & 22.0 & 27.4 & 43.4 & 47.2 & 75.2 & 83.6\\
Frequency~\cite{zellers2018neural} & 23.5 & 27.6 & 32.4 & 34.0 & 59.9 & 64.1 & 25.3 & 30.9 & 40.5 & 43.7 & 71.3 & 81.2\\
Frequency+Overlap~\cite{zellers2018neural} & 26.2 & 30.1 & 32.3 & 32.9 & 60.6 & 62.2 & 28.6 & 34.4 & 39.0 & 43.4 & 75.7 & 82.9\\
MotifNet-LeftRight~\cite{zellers2018neural} & 27.2 & 30.3 & 35.8 & 36.5 & 65.2 & 67.1 & 30.5 & 35.8 & \textbf{44.5} & 47.7 & 81.1 & 88.3\\
GraphRCNN~\cite{yang2018graph} & 11.4 & 13.7 & 29.6 & 31.6 & 54.2 & 59.1 & - & - & - & - & - & - \\
KERN~\cite{chen2019knowledge} & 27.1 & 29.8 & \textbf{36.7} & 37.4 & 65.8 & 67.6 & - & - & - & - & - & -\\
VCTREE~\cite{tang2019learning} & 27.7 & 31.1 & 37.9 & \textbf{38.6} & 66.2 & 67.9 & - & - & - & - & - & -\\
\hline
HOSE-Net ($K=40$) & \textbf{28.9} & \textbf{33.3} & 36.3 & 37.4 & \textbf{66.7} & \textbf{69.2} & \textbf{30.5} & \textbf{36.3} & 44.2 & \textbf{48.1} & \textbf{81.1} & \textbf{89.2}\\
\hline
\hline
RelDN$^\ast$~\cite{zhang2019graphical} & 28.3 & 32.7 & 36.8 & 36.8 & 68.4 & 68.4 & 30.4 & \textbf{36.7} & 48.9 & 50.8 & 93.8 & 97.8\\
HOSE-Net$^\ast$ ($K=40$) & \textbf{28.9} & \textbf{33.3} & \textbf{37.3} & \textbf{37.3} & \textbf{70.1} & \textbf{70.1} & \textbf{30.5} & 36.3 & \textbf{49.7} & \textbf{51.2} & \textbf{94.6} & \textbf{98.2}\\
\hline
\end{tabular}
\caption{Comparison with state-of-the-art methods on Visual Genome. HOSE-Net$^\ast$ uses the evaluation metric in ~\cite{zhang2019graphical}}
\label{tab:sota_vg}
\end{table*}

\subsection{Datasets}

\textbf{Visual Genome\cite{krishna2017visual}.} It is a large scale dataset with 75729 object classes and 40480 relation classes. There are several modified versions for scene graph generation. In this paper, we follow the same train/val splits in which the most frequent 150 objects and 50 relations are chosen. 

We measure our method on VG in three tasks:
\begin{enumerate}
    \item predicate classification (PRDCLS): Given the ground truth annotations of the object classes and bounding boxes, predict the relation type of each object pair.
    \item Scene graph classification (SGCLS):  Given the ground truth annotations of object bounding boxes, predict the object classes and the relation type of each object pair.
    \item Scene graph detection (SGDET): Predict the bounding boxes, the object classes and the relation type of each object pair.
\end{enumerate}

We use Recall@50, Recall@100 as our evaluation metrics. Recall@x computes the fraction of relationship hits in the top x confident relationship predictions. The reason why precision and average precision (AP) are not proper metrics for this task is, only a fraction of relationships are annotated and they will penalize the right detection if it is not in the ground truth.
We report the Graph Constraint Recall@x following \cite{lu2016visual} which only involves the highest score relation prediction of each subject-object pair in the recall ranking. We also report the No Graph Constraint Recall@x following \cite{newell2017pixels} which involves all the 50 relation scores of each subject-object pair in the recall ranking. It allows multiple relations exist between objects.


\noindent\textbf{VRD\cite{lu2016visual}} contains 4000 training and 1000 test images including 100 object classes and 70 relations.  

We follow \cite{yu2017visual} to measure our method on VRD in two tasks:
\begin{enumerate}
    \item Phase detection: Predict the visual phase triplets $\textless$subject-relation-object$\textgreater$ and localize the union bounding boxes of each object pair.
    \item Relationship detection: Predict the visual phase triplets $\textless$subject-relation-object$\textgreater$ and localize the bounding boxes of subjects and objects.
\end{enumerate}

We report Recall@50 and Recall@100 at involving 1 ,10 and 70 relation predictions per object pair in recall ranking as the evaluation metrics.

\begin{table*}
    \centering
    \resizebox{\linewidth}{!}{
    \begin{tabular}{l | l l l l l l | l l l l l l}
    \hline
     & \multicolumn{6}{c}{Relationship Detection} \vline & \multicolumn{6}{c}{Phrase Detection}\\
     & \multicolumn{2}{c}{rel=1} & \multicolumn{2}{c}{rel=10} & \multicolumn{2}{c}{rel=70} \vline& \multicolumn{2}{c}{rel=1} & \multicolumn{2}{c}{rel=10} & \multicolumn{2}{c}{rel=70}\\
     Recall at & 50 & 100 & 50 & 100 & 50 & 100 & 50 & 100 & 50 & 100 & 50 & 100\\
     \hline
     VTransE~\cite{zhang2017visual} & 19.4 & 22.4 & - & - & - & - & 14.1 & 15.2 & - & - & - & -\\
     ViP-CNN~\cite{li2017vip} & 17.32 & 21.01 & - & - & - & - & 22.78 & 27.91 & - & - & - & -\\
     VRL~\cite{liang2017deep} & 18.19 & 20.79 & - & - & - & - & 21.37 & 22.60 & - & - & - & -\\
     KL distilation~\cite{yu2017visual} & 19.17 & 21.34 & \textbf{22.56} & \textbf{29.89} & \textbf{22.68} & \textbf{31.89} & 23.14 & 24.03 & 26.47 & 29.76 & 26.32 & 29.43\\
     MF-URLN~\cite{zhan2019exploring} & 23.9 & 26.8 & - & - & - & - &  31.5 & 36.1 & - & - & - & -\\
     Zoom-Net~\cite{yin2018zoom} & 18.92 & 21.41 & - & - & 21.37 & 27.30 & 24.82 & 28.09 & - & - & 29.05 & 37.34\\
     CAI + SCA-M~\cite{yin2018zoom}      & 19.54 & 22.39 & - & - & 22.34 & 28.52 & 25.21 & 28.89 & - & - & \textbf{29.64} & \textbf{38.39}\\
     RelDN (ImageNet)~\cite{zhang2019graphical} & 19.82 & 22.96 & 21.52 & 26.38 & 21.52 & 26.38 & 26.37 & 31.42 & 28.24 & 35.44 & 28.24 & 35.44\\
     \hline
     HOSE-Net ($K=18$) & \textbf{20.46} & \textbf{23.57} & 22.13 & 27.36 & 22.13 & 27.36 & \textbf{27.04} & \textbf{31.71} & \textbf{28.89} & \textbf{36.16} & 28.89 & 36.16\\
     \hline
    \end{tabular}
    }
    \caption{Comparison with state-of-the-art methods on VRD.}
    \label{tab:sota_vrd}
\end{table*}

\subsection{Implementation Details}

HOSE-Net adopts a two-stage pipeline. The object detector is Faster RCNN with a VGG backbone initialized by COCO pre-trained weights for Visual Genome and ImageNet pre-trained weights for VRD and then finetuned on the visual relationship datasets. The backbone weights are fixed. 
For stable training, we add an unstructured relation classifier as a separate branch for joint training.
Considering the dataset scale and dataset quality, we adopt different training mechanisms for Visual Genome and VRD. In Visual Genome experiments, we set $lr = 0.001$ for the structured classifier and $lr = 0.01$ for the unstructured one. During testing, we evaluate the structured classifier. In VRD, the loss weight of the unstructured classifier is 0.7, and the structured one is 0.3. 
During testing, the result is the weight sum of the two classifiers.
Since the statistical bias is a widely-adopted strategy in the two-stage pipeline, we train a bias vector and fuse the bias results with the visual module results during testing following \cite{zellers2018neural}. 

The proposed framework is implemented by PyTorch. All experiments are conducted on servers with 8 NVIDIA Titan X GPUs with 12 GB memory.
The batch size at the training phase is set to 8. $d_f$ is set to 4096 and $d_e$ is set to 512.



\subsection{Ablation Study} Now we perform ablation studies to better examine the effectiveness of our framework.
\\
\textbf{Structured Output Space with Cluster Number K.} We perform an ablation study to validate the effectiveness of the SEC module which learns a structured output space and the HSA module which incorporates higher order structure into the output space with respect to the cluster number K.
$K = 1$ is our baseline model which uses the conventional unstructured relation classifiers.
$K = 150$ only employs SEC module to learn a low order structured output space.
In HSA module, K is a hyper parameter which can be a trade-off between the performance and the model complexity. 
We know that all clustering algorithms suffer from the lack of automatic decisions for an optimal number of clusters. While trying all possible combinations is prohibitively expensive, we have got a comprehensive set of results for comparison. The performance curve on $K=1,10,20,30,40,50,60,70,100,130,150$ are shown in the Figure~\ref{fig:compair_m}. 
We find $K=40$ works the best.
Table~\ref{tab:compair_m} presents results when $K=1,150,40$. \\ 
The comparison shows that:
\begin{enumerate}[1)]
    \item Adopting a structured output space ($K=40,150$) is superior to an unstructured one($K=1$) which verifies the effectiveness of the SEC module.
    \item Adopting a higher order structured output space ($K=40$) outperforms lower order one ($K=150$) which verifies the effectiveness of the HSA module.
\end{enumerate}
We also show the resulting semantic hierarchy of objects from the HSA module on VG and VRD in Figure~\ref{fig:optim_result}. Although the HSA module is not designed to sort out the objects, the unsupervised process of searching higher order connectivity patterns in the knowledge graph can contribute to an object taxonomy. At the lower levels, the object classes are classified according to more specific properties, eg. roof with railing, street with sidewalk, train with car. At the higher levels, the clusters have more abstract semantics and are classified according to more general properties, eg. glass-bottle-cup with basket-box-bag, toilet-sink-drawer with shelf-cabinet-counter. 

\noindent\textbf{Clustering in HSA}.
Our behavior pattern based hierarchical clustering purely relies on the knowledge graph structure of the ground truth.
To verify the effectiveness of our clustering, we also conduct K-means clustering on word2vec embeddings of objects to obtain an external knowledge based clusters.
Table~\ref{tab:kmeans} presents the results of adopting HSA clustering results ($K=40$), adopting K-means with word2vec embedding clustering results and not adopting context clustering $K=150$).\\
The comparison shows that:
\begin{enumerate}[1)]
    \item HOSE-Net with SEC($K=150$) shows comparable results to HOSE-Net with SEC and K-means with word2vec embedding($K=40$), which means, the clustering results can't improve the performance.\\
    \item HOSE-Net with SEC and HSA($K=40$) outperforms HOSE-Net with SEC and K-means with word2vec embedding($K=40$), which proves that our structure-based clustering with internal relation knowledge can truly produces helpful clustering results to boost this task.
\end{enumerate}


\subsection{Comparison to State of the Art} 
\textbf{Visual Genome:} Table~\ref{tab:sota_vg} shows the performance of our model outperforms the state-of-the-art methods. Our object detector is adopted from \cite{zhang2019graphical} with $mAP = 25.5$ , $IoU = 0.5$. The number of clusters for comparison is 40. These methods all adopt flat relation classifiers. VRD\cite{lu2016visual}, AsscEmbed\cite{newell2017pixels}, Frequency\cite{zellers2018neural} predict the objects and the relations without joint inference. The other works are engaged in modeling the inter-dependencies among objects and relations. MotifNet-LeftRight\cite{zellers2018neural} encodes the dependencies through bidirectional LSTMs. MSDN\cite{li2017scene},ISGG\cite{xu2017scene},KERN\cite{chen2019knowledge} rely on message passing mechanism. SGP\cite{herzig2018mapping} employs structured learning.  In comparison, our framework doesn't refine the visual representations but still achieves new state-of-the-art results on SGDET, SGCLS, PRDCLS with and without graph constraint.
RelDN\cite{zhang2019graphical} proposes contrastive losses and reports Top@K Accuracy (A@K) on PredCls and SGCls in which the ground-truth subject-object pair information is also given.
We also compare with RelDN at A@K as shown in Table~\ref{tab:sota_vg}.

\noindent\textbf{VRD:} Table~\ref{tab:sota_vrd} presents results on VRD compared with state-of-the-art methods. The number of clusters for comparison is 18. The implementation details of most methods on VRD are not very clear. As shown in \cite{zhang2019graphical}, pre-training on COCO can provide stronger localization features than pre-training on ImageNet. For a fair comparison, we use the ImageNet pre-trained model. We achieve new state-of-the-art results on Relationship Detection and Phrase Detection.

\section{Conclusions}
In this work, we propose Higher Order Structure Embedded Network to address the problems caused by ignoring the structure nature of scene graphs in existing methods.
First we propose a Structure-Aware Embedding-to-Classifier module to redistribute the training samples into different classification subspaces according to the object labels and connect the subspaces with a set of context embeddings following the global knowledge graph structure. 
Then we propose a Hierarchical Semantic Aggregation module to mine higher order structures of the global knowledge graph which makes the model more scalable and trainable.

\section{ACKNOWLEDGMENTS}
This work was supported by NSFC project Grant No. U1833101, SZSTI under Grant No. JCYJ20190809172201639 and the Joint Research Center of Tencent and Tsinghua.

\bibliographystyle{ACM-Reference-Format}
\bibliography{submit}

\end{document}